# Biomedical Knowledge Graph Refinement and Completion using Graph Representation Learning and Top-K Similarity Measure

Islam Akef Ebeid[1], Majdi Hassan[2], Tingyi Wanyan[3], Jack Roper[1], Abhik Seal[2], Ying Ding[1]

The University of Texas at Austin[1], AbbVie Inc.[2], Indiana University[3]

**Abstract.** Knowledge Graphs have been one of the fundamental methods for integrating heterogeneous data sources. Integrating heterogeneous data sources is crucial, especially in the biomedical domain, where central data-driven tasks such as drug discovery rely on incorporating information from different biomedical databases. These databases contain various biological entities and relations such as proteins (PDB), genes (Gene Ontology), drugs (DrugBank), diseases (DDB), and protein-protein interactions (BioGRID). The process of semantically integrating heterogeneous biomedical databases is often riddled with imperfections. The quality of data-driven drug discovery relies on the accuracy of the mining methods used and the data's quality as well. Thus, having complete and refined biomedical knowledge graphs is central to achieving more accurate drug discovery outcomes. Here we propose using the latest graph representation learning and embedding models to refine and complete biomedical knowledge graphs. This preliminary work demonstrates learning discrete representations of the integrated biomedical knowledge graph Chem2Bio2RD [3]. We perform a knowledge graph completion and refinement task using a simple top-K cosine similarity measure between the learned embedding vectors to predict missing links between drugs and targets present in the data. We show that this simple procedure can be used alternatively to binary classifiers in link prediction.

## 1 Introduction

The Knowledge Graph provides a unique opportunity as an emerging model to break data and information silos through effective data integration techniques and methods [11]. The term Knowledge Graph (KG) has emerged to describe the technology that the Google search engine started using in 2012 [16]. A KG is a graph representation of interrelated and semantically connected entities [11]. KG construction can vary from manual curation to automatic construction via extracting entities and relations from unstructured text using Natural Language Processing [16]. KGs can also be constructed from existing heterogeneous relational databases through integration methods [3]. KGs can be represented using the property graph model [11] described as a list of edges between unique nodes and properties or using the Resource Descriptor Framework (RDF) format defined by the World Wide Web Consortium [12]. RDF is a graph model that uses a markup language similar to XML to represent data. RDF represents knowledge graphs as triples of a head entity, a tail entity, and a relationship type. The RDF triples can be used to model facts in the form of first-order logic. For example, <Mount Fuji, isLocatedIn, Japan> where <Mount Fuji> and <Japan> are two unique nodes or entities in the graph, and <isLocatedIn> is the relationship between the two nodes.

Despite the multitude of methods used in constructing and curating knowledge graphs, they are susceptible to incomplete and inaccurate data [16]. Knowledge Graph Refinement and Completion aim to improve KGs by filling in its missing knowledge and checking the existing knowledge's validity. Here we propose using graph representation learning models to learn different discrete feature representations of entities in Chem2Bio2RDF. We use vector operations such as cosine distance as a similarity ranking measure to predict missing knowledge and links between drugs and potential targets [5] to complete and refine the knowledge graph.

## 2    Background

In [3], Chem2Bio2RDF was introduced as an integrated biomedical knowledge graph. In [5], the authors suggested using meta path-based sampling from Chem2Bio2RDF to extract features that would be used to test multiple machine learning models trained to predict missing drug-target interactions within the KG. Revealing possible side effects of specific drug and gene interactions is a highly sought goal in data-driven drug discovery and chemical systems biology [21]. Predicting drug-target interactions lends itself to the task of link prediction in graph analytics [5]. Link prediction on biomedical knowledge graphs aims to identify potential associations between entities such as drugs, proteins, and genes, a task known as knowledge graph completion and refinement.

Representation learning aims at learning latent feature vectors from data without relying on stochastic and heuristic metrics and measures [1]. In graphs, all representation learning algorithms ultimately produce node embedding vectors in a low-dimensional vector space. The minimal constraint on the learned representations is that they preserve the graph's structure in the Euclidean vector space [8]. The node embedding vectors learned can be passed to downstream machine learning models for classification, regression, or clustering. Alternatively, they can be used directly in the learning process in a semi-supervised end to end fashion as in Graph Neural Networks Approaches [20]. They can also be inductive as in learned from the structure of the graph itself [17] [18] [7] or transductive by forcing a scoring function to evaluate the plausibility of the triples in the KG [14] [2].

Link prediction-based knowledge graph completion using representation learning relies mostly on binary classifiers with the task of learning a model that would predict whether a link should exist between two entities on a subset of the graph [13]. We argue that binary classifiers for link prediction based biomedical KG completion are unnecessary [10]. Instead, using similarity functions between learned node embedding vectors is sufficient to predict missing links between bioentities with more realistic uncertainty. We evaluate our models using an experimental ground truth dataset collected from the DrugBank [22] on positive and negative associations between potential drugs and targets such as genes, proteins, and chemical ontologies.

## 3   Method

The class of network embedding models used in this paper relies on random walk sampling of the input graph followed by an unsupervised Skip-gram model described initially [15]. Different network embedding algorithms learn different embedding vectors for each node. That largely depends on the information each algorithm tries to preserve during learning. In node2vec [7], structural information is preserved, while in edge2vec [6], structural plus edge types are preserved. In metapath2vec [4], node types, relations, and topology are preserved. While in GraphSAGE [9], the structure or the topology is preserved, yet embeddings for out of sample nodes can be inferred.

### 3.1   The Skip-gram Model

The Skip-gram model described in [15] is an unsupervised language model aiming at learning discrete vectors of unique words given a corpus of text. In unsupervised network and graph embedding algorithms, the Skip-gram model has been adapted to learn vectors for individual nodes on a corpus of sampled nodes from the graph. In DeepWalk [17], the authors first proposed using the Skip-gram model on a corpus of sampled nodes from a given graph using a random walk strategy. Since it was observed that the distribution of words in any corpus of text followed Zipf's law, it was also observed that the frequency distribution of nodes in a graph follows Zipf's law [17]. The random walk algorithm aims at sampling chains of nodes from the graph controlled by the walk length. Sampled chains are then treated as a text corpus fed to a Skip-gram variant of Word2vec. In Word2Vec, the text is first tokenized into sentences; similarly, the random walks mimic the process of creating meaningful "sentences" out of the graph as a sampling approach.

The Skip-gram model can be formalized given a sequence of training words or nodes $(w_1, w_2, w_3, \ldots, w_t)$ the goal is then to maximize the log-likelihood probability of the current word given the surrounding words:

$$P(w_1, w_2, w_3, \ldots, w_t) = \frac{1}{T}\sum_{T}^{t=1} \sum_{-c \leq j \leq c, j \neq 0} \log P(w_{t+j}|w_t) \qquad (1)$$

C is the size of the training window around the current word, and T is the size of the training corpus. Computing $P(w_{t+j}|w_t)$ requires computing a full Softmax function, which is intractable. Instead, a sampling approach, such as Hierarchical Softmax or Negative Sampling, is needed, as described in [15]. Negative sampling is mostly used in network embedding. It is done by reformulating the neural network's output layer to become a logistic regression-based classifier rather than a Softmax predictor. The input is a context window of words where each word is classified as 1 within the context window. Otherwise, 0 if the word was negatively sampled from outside of the context window. A Stochastic Gradient Descent [23] is then used to optimize the objective function to learn the feature vector U for each unique word in the vocab V as showing in eq (2). Note that in the case of network embedding for a graph G = (N, E) W becomes N, where words become nodes. The input corpus to the Skip-gram model becomes the output corpus of sampled nodes using the random walk strategy.

$$J_t(\theta) = \log \sigma \left(u_o^T v_c\right) + \sum_{j=P(w)} \left[\log \sigma \left(-u_j^T v_c\right)\right] \qquad (2)$$

**Node2vec** [7] performs a modified version of the random walk where it includes parameters p and q to control the sampling strategy. The p parameter controls the likelihood of the walk revisiting a node that has already been visited. The q parameter controls whether the search is constrained locally or globally. Given q > 1 and a random walk on an initial node, the random walk samples nodes closer to the initial node as in Breadth-First Search. Whereas q < 1, random walk samples nodes further from the initial node similar to a Depth First Search. This customizability in search behavior allows the walk to capture diverse structural properties within the graph. The sampling strategy builds a corpus for each node. A Skip-gram model trains on this corpus to generate a unique embedding vector for each node.

**Edge2vec** [6] builds on Node2vec by introducing an additional sampling step where edge semantics are considered during the random walk sampling. The random walker first trains a transition matrix built from the edges using an Expectation-Minimization algorithm. A corpus of nodes is generated from random walks that consider the edge weights from the transition matrix. Then, a Skip-gram train on the corpus to generate embeddings.

In **Metapath2vec** [4], the random walk sampling strategy is guided by predefined meta paths. KGs are usually defined by a schema, as in figure 1. A meta path is determined from the KG schema that generally has a semantic meaning within the KG domain knowledge. For example, the meta path protein-protein-gene refers to a biochemical process of protein interactions and genes binding to those proteins, reflecting a semantic meaning. The sampled corpus of nodes is then passed to a Skip-gram model to learn the embedding vectors.

**GraphSAGE** [9] is an inductive GNN based approach to solving the problem of what if after we trained an embedding model for the existing graph, a new node comes in?

Previous node embedding algorithms rely on a transductive approach to generate node embedding by training a model that would leverage the graph structure and topology information. GraphSAGE offers a way to leverage the already existing node embedding to produce embedding feature vectors for newly introduced nodes inductively. That is done by aligning the new observed nodes to the existing trained embedding space using aggregation and pooling functions on the target node's neighborhood.

### 3.2 Dataset

Advances in biomedical sciences have given rise to new integrated and significant biological data sources. Integrating these data sources involves converting the graph to RDF format and then integrating them semantically. Chem2Bio2RDF [3] is an extended version of Bio2RDF [24]. It aggregates data from multiple sources from Bio2RDF [24], DrugBank [22], and PubChem [25]. Chem2Bio2RDF contains over 700000 triples and 295000 unique entities. A schema of Chem2Bio2RDF is presented in figure 1, showing 9 bioentities and their semantic relationships in the graph.

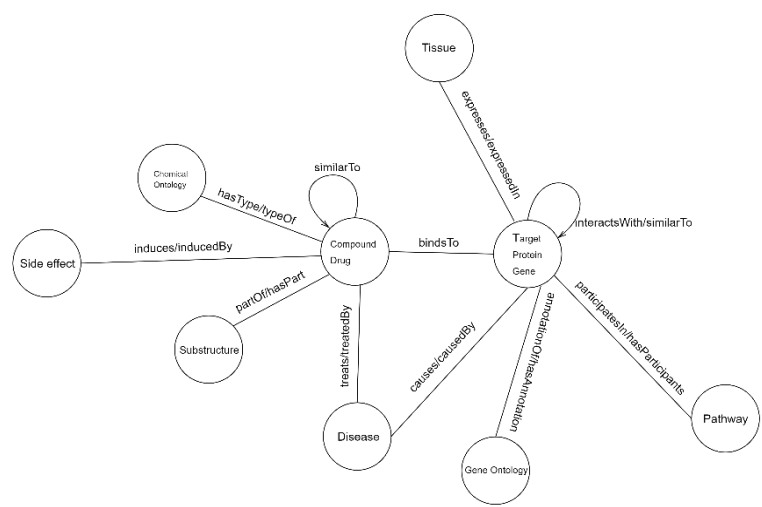

**Fig. 1.** The schema describes entities and relationships in Chem2Bio2RDF. For example, the graph can contain a triple that describes a relationship between a compound/drug as part of a substructure and a substructure that is part of a compound/drug.

### 3.3 Model Training

The model's input is a list of Chem2Bio2RDF triples where the source node and the target edge describe the relationship. The graph is then sampled to build a corpus of node walks, which can then be used as an input to the Skip-gram algorithm. Sampling strategies differ based on the algorithm used, as described before in section 3.1. For example, in Metapath2vec, the walk length we used was 20, and the number of walks per node was 10. Vector dimensions for node2vec, metapath2vec, and edge2vec were chosen to be 128, while for GraphSAGE was 500. The metapaths used in training metpath2vec are described in [16].

## 4    Experiments and Results

Predicting non-existing relationships between entities in the KG is known as Knowledge Graph Completion [16]. We can predict possible relationships between different entities that have not been explored before, speeding up discovering new applications of existing drugs [5]. We focus primarily on completing relationships between drugs and targets in Chem2Bio2RDF. We pose drug-target interaction prediction as top-k similarity ranking [26]. We used drug-target interaction data as ground truth extracted from DrugBank [22] to evaluate link prediction models as the top-k similarity between node embedding vectors. A cosine distance similarity metric is used to compute a similarity score between the target node vector and the original node vector.

$$\text{similarity} = \cos(\theta) = \frac{AB}{||A||\,||B||} \quad (3)$$

Across these tasks, accuracy, precision, recall, and F1 scores were recorded in tables 2, 3, and 4. Table 1 shows the results from a link prediction task using a logistic regress binary classifier. Would a simple top-K similarity score between vectors be sufficient to assess the suitability of predicting new links between drugs and targets in the graph instead of using binary classifiers?

We evaluate the performance of the 4 algorithms on a test set of possible and impossible links between drugs and target using two methods the Top-K similarity method [26] and the binary classification based link prediction method [10]. We use two test sets containing a set of drug-target pairs that include positive and negative labels for a link between a pair. The first test set was used in [5] consists of 5,387 positive pairings and 26,682 negative pairings. The second test set extracted from the DrugBank[22] contains drug-target interaction pairs that do not exist in Chem2Bio2RDF but should exist. That test set has 5,836 positive pairings and 2,368 negative pairings. For both test sets, The positive and negative pairings were verified experimentally through ChEMBL [27], and the negative pairings had at least 5 uM of activity. For both evaluation methods, we used the 2 test sets to evaluate the performance of the 4 algorithms.

We trained a logistic regression model on labeled positive and negative links sampled from the binary classification-based link prediction graph. The model was then tested on a validation set sampled from the graph.

Like the binary classification based link prediction task, top-k ranking predicts whether a link exists between two entities. Unlike link prediction, where it provides a single prediction between a compound and gene, the top-K ranking task considers drug pairing to several targets and ranks based on the cosine similarity scores between their learned vectors. That places the importance on the rank of the drug-target link rather than the existence of a link. The ranking provides a better probability of a link between two entities; however, it comes at the cost of efficiency as it is more computationally expensive. After the algorithm ranks the similarity scores between two embedding vectors, we evaluate the top-K portion of the rankings.

The accuracy results are close and similar between both the binary classifier method and the top-K method. That supports our argument that top-K similarity prediction can be used for Knowledge Graph Completion via link prediction. Despite the variability in the f1 score, accuracy remains very close between the two methods, suggesting that training better models using graph representation learning algorithms would provide par performance. Besides, link prediction using binary classification models provides a rigid framework incapable of capturing the uncertainty in graph representation learning for knowledge graph completion.

**Table 1.** The results for link prediction using logistic regression

|  | **Accuracy** | **F1** | **Precision** | **Recall** |
|---|---|---|---|---|
| **Node2vec** | 0.9920 | 0.9631 | 0.9843 | 0.9427 |
| **Edge2vec** | 0.9922 | 0.9637 | 0.9873 | 0.9412 |
| **Metapath2vec** | 0.9935 | 0.9700 | 0.9890 | 0.9517 |
| **GraphSAGE** | 0.9865 | 0.9366 | 0.9739 | 0.9019 |

**Table 2.** The results of Top-K similarity where K=10

|  | **Accuracy** | **F1** | **Precision** | **Recall** |
|---|---|---|---|---|
| **Node2vec** | 0.9314 | 0.5838 | 0.8838 | 0.4359 |
| **Edge2vec** | 0.9567 | 0.7830 | 0.8766 | 0.7074 |
| **Metapath2vec** | 0.9611 | 0.8132 | 0.8682 | 0.7647 |
| **GraphSAGE** | 0.8880 | 0.0260 | 0.3214 | 0.0136 |

**Table 3.** The results of Top-K similarity where K=50

|  | **Accuracy** | **F1** | **Precision** | **Recall** |
|---|---|---|---|---|
| **Node2vec** | 0.9421 | 0.7005 | 0.8156 | 0.6139 |
| **Edge2vec** | 0.9596 | 0.8209 | 0.8026 | 0.8401 |
| **Metapath2vec** | 0.9590 | 0.8237 | 0.7844 | 0.8673 |
| **GraphSAGE** | 0.8834 | 0.0514 | 0.2500 | 0.0287 |

**Table 4.** The results of Top-K similarity where K=100

|  | Accuracy | F1 | Precision | Recall |
|---|---|---|---|---|
| **Node2vec** | 0.9438 | 0.7317 | 0.7722 | 0.6953 |
| **Edge2vec** | 0.9514 | 0.7966 | 0.7400 | 0.8627 |
| **Metapath2vec** | 0.9476 | 0.7881 | 0.7131 | 0.8808 |
| **GraphSAGE** | 0.8787 | 0.0853 | 0.2537 | 0.0528 |

### 4.1 Use Case:

Table 5 shows the data available in Chem2Bio2RDF for the drug Apicidin (467801). Using a simple top-20 similarity query over our trained Metapath2vec model shown in table 6, we predicted that the drug Apicidin should be linked to the gene HDA106. Since it has a higher cosine similarity score than some of the genes that already exist in the graph. Apicidin is a histone deacetylase inhibitor used to treat tumors [19]. HDA106 and all the HAD genes are critical enzymes involved in developing cancer and other diseases such as interstitial fibrosis, autoimmune, inflammatory diseases, and metabolic disorders [19]. That suggests that our procedure could predict that a link should exist between Apicidin and HDA106 in Chem2Bio2RDF, even though that link did not exist before. The link between Apicidin and HAD106 was also not predicted using the binary classifier link predictor.

**Table 5.** The links in the graph for the drug Apicidin.

| Source Node | Target Node |
|---|---|
| http://chem2bio2rdf.org/pubchem/resource/pubchem_compound/467801 | http://chem2bio2rdf.org/uniprot/resource/gene/HDAC5 |
| http://chem2bio2rdf.org/pubchem/resource/pubchem_compound/467801 | http://chem2bio2rdf.org/uniprot/resource/gene/HDAC6 |
| http://chem2bio2rdf.org/pubchem/resource/pubchem_compound/467801 | http://chem2bio2rdf.org/uniprot/resource/gene/HDAC10 |
| http://chem2bio2rdf.org/pubchem/resource/pubchem_compound/467801 | http://chem2bio2rdf.org/uniprot/resource/gene/HDAH |
| http://chem2bio2rdf.org/pubchem/resource/pubchem_compound/467801 | http://chem2bio2rdf.org/uniprot/resource/gene/HDAC4 |
| http://chem2bio2rdf.org/pubchem/resource/pubchem_compound/467801 | http://chem2bio2rdf.org/uniprot/resource/gene/HDAC7 |
| http://chem2bio2rdf.org/pubchem/resource/pubchem_compound/467801 | http://chem2bio2rdf.org/uniprot/resource/gene/NCOR2 |
| http://chem2bio2rdf.org/pubchem/resource/pubchem_compound/467801 | http://chem2bio2rdf.org/uniprot/resource/gene/HDAC11 |
| http://chem2bio2rdf.org/pubchem/resource/pubchem_compound/467801 | http://chem2bio2rdf.org/uniprot/resource/gene/F3 |
| http://chem2bio2rdf.org/pubchem/resource/pu | http://chem2bio2rdf.org/uniprot/res |
| bchem_compound/467801 | ource/gene/HDAC1 |
| http://chem2bio2rdf.org/pubchem/resource/pubchem_compound/467801 | http://chem2bio2rdf.org/uniprot/resource/gene/HDAC9 |
| http://chem2bio2rdf.org/pubchem/resource/pubchem_compound/467801 | http://chem2bio2rdf.org/uniprot/resource/gene/HDAC8 |
| http://chem2bio2rdf.org/pubchem/resource/pubchem_compound/467801 | http://chem2bio2rdf.org/uniprot/resource/gene/HDAC2 |
| http://chem2bio2rdf.org/pubchem/resource/pubchem_compound/467801 | http://chem2bio2rdf.org/uniprot/resource/gene/HDAC3 |

**Table 6.** All top 20 are target genes/proteins for the drug Apicidin (467801). The higher the cosine distance score, the more probable that the link between them should exist in the graph.

| Gene | Cosine Distance Score |
|---|---|
| http://chem2bio2rdf.org/uniprot/resource/gene/NCOR2 | 0.8363208028622343 |
| http://chem2bio2rdf.org/uniprot/resource/gene/HDAC10 | 0.7164958960454791 |
| http://chem2bio2rdf.org/uniprot/resource/gene/HDAC11 | 0.6595226647235345 |
| http://chem2bio2rdf.org/uniprot/resource/gene/HDAC7 | 0.6594273898963945 |
| http://chem2bio2rdf.org/uniprot/resource/gene/HDAC6 | 0.6459631340208143 |
| http://chem2bio2rdf.org/uniprot/resource/gene/HDAC9 | 0.6364544696270995 |
| http://chem2bio2rdf.org/uniprot/resource/gene/HDAH | 0.6329007338447462 |
| http://chem2bio2rdf.org/uniprot/resource/gene/HDAC4 | 0.6079327615804068 |
| http://chem2bio2rdf.org/uniprot/resource/gene/HDAC3 | 0.6070127461182808 |
| http://chem2bio2rdf.org/uniprot/resource/gene/HDA106 | 0.6047486633717029 |
| http://chem2bio2rdf.org/uniprot/resource/gene/HDAC5 | 0.6029446879584531 |
| http://chem2bio2rdf.org/uniprot/resource/gene/HDAC1 | 0.5987907960296356 |
| http://chem2bio2rdf.org/uniprot/resource/gene/HDAC2 | 0.5682675274176112 |
| http://chem2bio2rdf.org/uniprot/resource/gene/HD1B | 0.5497739534316449 |
| http://chem2bio2rdf.org/uniprot/resource/gene/HDAC8 | 0.5227708345535201 |
| http://chem2bio2rdf.org/uniprot/resource/gene/ACUC1 | 0.4979581350083667 |
| http://chem2bio2rdf.org/uniprot/resource/gene/BCOR | 0.4433902007677988 |
| http://chem2bio2rdf.org/uniprot/resource/gene/BTBD14B | 0.440922126476731243 |
| http://chem2bio2rdf.org/uniprot/resource/gene/PHF21A | 0.4381844427483992 |
| http://chem2bio2rdf.org/uniprot/resource/gene/FLI1 | 0.4370752763874226 |

## 5   Conclusion

We demonstrated graph representation learning for Knowledge Graph Completion using Top-K similarity distance between the learned embeddings in this preliminary work. The results of the model need to be further evaluated to be included in the dataset itself. Less semantically aware algorithms performed better than mode semantically aware algorithms in link prediction. That might be because less semantically aware algorithms are sufficient to predict direct and simple links like drugs and genes.

While more semantically aware algorithms might be better for predicting relationships between more complicated pathways. Future work includes investigating incorporating more semantics in embedding models and testing whether these heterogeneous models perform better than a homogenous model. In the future, we also intend to include a full parameter sensitivity study alongside an extended version of the evaluation protocol. We need to experiment on the feasibility of reliably using top-K similarity for link prediction for knowledge graph completion for drug discovery.